\documentclass[a4paper,conference]{IEEEtran}
\usepackage{graphicx}
\usepackage{eurosym}
\usepackage{cite}
\usepackage{amsmath}
\usepackage{algorithmic}
\usepackage{array}
\usepackage{url}
\usepackage{arydshln}
\usepackage{color}
\newcommand\blfootnote[1]{%
  \begingroup
  \renewcommand\thefootnote{}\footnote{#1}%
  \addtocounter{footnote}{-1}%
  \endgroup
}
\newcommand{\comment}[1]{}
\begin{document}
%
% paper title
% Titles are generally capitalized except for words such as a, an, and, as,
% at, but, by, for, in, nor, of, on, or, the, to and up, which are usually
% not capitalized unless they are the first or last word of the title.
% Linebreaks \\ can be used within to get better formatting as desired.
% Do not put math or special symbols in the title.
\title{A Simple Domain Shifting Network \\
for Generating Low Quality Images}

% author names and affiliations
% use a multiple column layout for up to three different
% affiliations
\author{%
\IEEEauthorblockN{Guruprasad Hegde$^*$, Avinash Nittur Ramesh$^*$, Kanchana Vaishnavi Gandikota$^*$, \\
Roman Obermaisser, Michael Moeller}
\IEEEauthorblockA{Department for Computer Science and Electrical Engineering\\
University of Siegen \\
\{guruprasad.hegde, avinash.ramesh, kanchana.gandikota\}@student.uni-siegen.de,\\ \{roman.obermaisser, michael.moeller\}@uni-siegen.de}
}
\maketitle

\blfootnote{$^*$equal contribution}

\begin{abstract}
Deep Learning systems have proven to be extremely successful for image recognition tasks for which significant amounts of training data is available, e.g., on the famous ImageNet dataset. We demonstrate that for robotics applications with cheap camera equipment, the low image quality, however, influences the classification accuracy, and freely available data bases cannot be exploited in a straight forward way to train classifiers to be used on a robot. As a solution we propose to train a network on degrading the quality images in order to mimic specific low quality imaging systems. Numerical experiments demonstrate that classification networks trained by using images produced by our quality degrading network along with the high quality images outperform classification networks trained only on high quality data when used on a real robot system, while being significantly easier to use than competing zero-shot domain adaptation techniques. 
\end{abstract}

\IEEEpeerreviewmaketitle

\newcommand{\MM}[1]{{\color{red} #1}}
\section{Introduction}
On a closed set of images with predefined classes and controlled conditions, recent machine learning approaches match or even surpass human image classification abilities. Challenging situations, however, arise when transferring such computer vision systems into real world practical applications, in which the distribution and characteristics of the images differs from the distribution of the online training examples significantly.

\begin{figure}
    \centering
    \small
    \begin{tabular}{lll}
        \includegraphics[width=2.5cm, height=2.5cm]{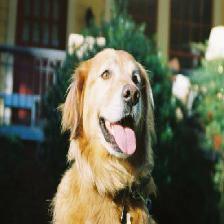}
        & \includegraphics[width=2.5cm, height=2.5cm]{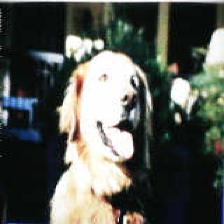} 
        & \includegraphics[width=2.5cm, height=2.5cm]{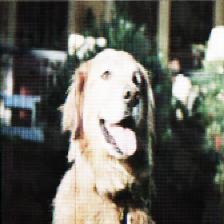} \\
        Standard: 1.0 dog 
        & Standard: 0.007 dog
        & Standard: 0.038 dog \\
        Adapted: 1.0 dog 
        & Adapted: 1.0 dog
        & Adapted: 1.0 dog \\
        %\includegraphics[width=2.5cm,height=2.5cm]{dog.2_Stanford.jpg}
        %\includegraphics[width=2.5cm, height=2.5cm]{dog.2_Cozmo}
        %\includegraphics[width=2.5cm, height=2.5cm]{dog.2_AE}
        %(a) Original  & (b) Degraded & Cozmo 
    \end{tabular}
    \caption{Images from online databases as illustrated on the left are often of decent quality and standard networks  (here: Standard \cite{MobileNetV2}) trained on such good quality images achieve high classification accuracy. Low cost cameras, however, introduce  distortions (middle image) in terms of the dynamic range and color reproduction, which can easily mislead standard networks. With the proposed technique of training classifiers on images that have been send through a quality degrading network to mimic low-cost cameras (right image), one obtains an adapted network that is significantly more robust with respect to low-quality images as illustrated in the classification scores below the respective images.}
    \vspace*{-5mm}
    \label{fig:teaser}
\end{figure}

As an example, we consider the problem of image classification in a video stream recorded with an Anki Cozmo$^{\copyright}$ robot camera. As the robot is only about $4\times 3 \times 2$ inches, and currently costs about $\text{\euro}\; 100$ only, the quality of recorded images is rather low compared to typical image classification data sets. As exemplified in Fig. \ref{fig:teaser} and detailed in Section~\ref{sec:results}, the specific distortion in terms of the dynamic range, color reproduction, and noise causes the accuracy of image classification networks trained on usual online data sets to drop significantly. The latter implies that standard benchmark datasets cannot be harvested directly in order to train networks for such devices on solving various vision based tasks.

 A large variety of different works have considered domain adaption methods (based on the availability of different types of examples from the source and target domain) for such situations. Their ideas include fine-tuning on labeled data of the new domain using adversarial training schemes to encourage a confusion between domains, or combining the classification with a reconstruction of data from the source domain. We refer to \cite{survey1} and Section \ref{sec:related} for details.
\graphicspath{ {./Images/} }
\vspace{.25cm}

In this work, our objective is to classify low quality images captured by robot camera by leveraging high quality labeled data without any low quality training data specific to the classification task. However, we assume that unlabeled high quality-low quality image pairs are available which are unrelated to the final task of interest~(classification). For this challenge, we propose a very simple, yet generic and powerful solution: We propose to train a simple convolutional regression network to degrade high quality images in such a way that their appearance mimics that of the same image recorded with a specific low quality camera. Subsequently, we use this domain shifting neural network to transfer any online classification dataset to a corresponding low-quality version of it and demonstrate in several numerical experiments that classifiers trained/fine-tuned on the low-quality versions yield better accuracy on images actually recorded with a low quality camera, see illustration in Fig. \ref{fig:teaser}. 
%Though this is straightforward, we are not aware of any existinnote that using a simple image-to-image mapping network to aid domain adaptation is novel and has not been considered in the existing literature. 
To generate a training set for the domain shifting neural network, we record high quality images displayed on a screen with the given low-cost camera  (see Fig.~\ref{fig:recording}) and train the network to map the high quality images to their corresponding low quality counterparts.  Since the images used to train the quality degrading network differ significantly from the images the classifier is trained on, the proposed approach can serve as a simple and generic scheme for adapting networks for various different tasks to the specific characteristics of a given low-cost camera.

\section{Related Work}
\label{sec:related}
Within the last years deep classification network architectures have been adapted to work well on mobile systems with limited computation power, with MobileNet~\cite{MobileNetV2} and SqueezeNet~\cite{SqueezeNet} being among the most popular variants. Such networks achieve accuracies close to their significantly more computationally expensive relatives on popular computer vision benchmarks, but the architectural considerations do not specifically account for low quality input data. 

Despite the continuously growing amount of labeled training data from various sources, care has to be taken of adapting any image classification network trained on online data bases to the specific setting it is meant to operate in. While it is common practice to encourage certain invariances, e.g. with respect to small noise, rotations and scale using data augmentations, it has been observed that the image quality can have a significant impact on the classification result (see e.g. \cite{donahue2014decaf,dodge2016understanding}).% such that adapting classification networks to the specific characteristics of their input images is of utmost importance. 

The problem of adapting a trained network to a new distribution of input images is a well-studied problem under the name of domain adaptation, see e.g. \cite{survey1} for a comprehensive overview. Divergence-based domain adaptation approaches~\cite{sun2016deep, rozantsev2018beyond, lee2019sliced} obtain domain invariant data representations by minimizing some divergence measure between source and target distributions.  Another approach is to employ adversarial training~\cite{liu2016coupled, yoo2016pixel} to encourage a confusion between source and target domains. Alternately, a common representation for source and target domains can be constructed by combining classification with an auxiliary reconstruction task~\cite{ghifary2016deep}.

In terms of the available training data, one can distinguish 4 categories of domain adaptation techniques. In the \textit{supervised} case, one has a well-trained network on a set of source domain images with given labels, along with a reduced number of labeled images from a target domain, and when sufficiently many labeled images in target domain are available,  simple fine-tuning/transfer learning techniques~\cite{yosinski2014transferable,chu2016best} can be applied. In \textit{semi-supervised domain adaptation}, weaker information is available in the target domain only, e.g. the works~\cite{Yao_2015_CVPR, ao2017fast, Saito_2019_ICCV} utilize additional unlabeled data for transferring knowledge to target domain when few  labeled  examples  are  available in the target domain. 
\textit{Unsupervised domain adaptation} techniques~\cite{huang2018domain, Roy_2019_CVPR, pan2019transferrable} address the scenario where a network pretrained on labeled source domain data is adapted to a target domain of unlabeled images that are related to the source domain, e.g. containing the same classes/objects. Finally, having no labeled images in the target domain for the task of interest  refers to \textit{zero-shot domain adaptation}, see e.g. ~\cite{peng2018zero, kumagai2018zero, pmlr-v101-ishii19a, Wang_2019_ICCV}. 

Our work can be seen as a specific form of zero-shot domain adaptation, as we do not consider to have labeled target data and usually not even target images that show the same classes as the classifier we'd like to train. By recording images from a screen we do, however, ensure we have a one-to-one correspondence of images from the source and target domain. These corresponding images are, however, entirely unlabeled. 

In contrast to our assumptions,~\cite{pmlr-v101-ishii19a} and~\cite{kumagai2018zero} do not have target domain data unrelated to the task of interest at all. Ishii~\emph{et al.}~\cite{pmlr-v101-ishii19a} assume good prior knowledge about attributes causing  distribution shift between source and target domains, which is used in adapting to the target domain. Kumagai and Iwata~\cite{kumagai2018zero} present the concept of latent domain vectors to represent multiple source domains, which are then used to find models for unseen target domains via bayesian inference. Similar to our assumptions,~\cite{Wang_2019_ICCV,peng2018zero} have  target domain data unrelated to the task of interest. However, Wang and Jiang~\cite{Wang_2019_ICCV}  do not assume correspondences between the source and target domain samples unrelated to the task of interest. They instead employ two generative adversarial networks~(GANs) to learn the joint distribution of source and target domain data across two tasks. The work~\cite{peng2018zero}  is closest to the scenario we considered in this work. Akin to us, their approach assumes that paired data in source and target domains for an irrelevant task are available. Their approach involves two steps: first matching features of the target domain irrelevant images with the features of the source domain images from a pretrained source domain network. Second, training the source network on the relevant task, while  maintaining feature similarity with the target network on the irrelevant task.  The important difference, however, is that~\cite{peng2018zero} assumes to have labeled paired data for classification of images (dissimilar to those considered in final classification). 

Since we assume that only unlabeled source-target image pairs are available, we train a simple regression network to map from high quality source images to low quality target images. However, our generic approach can also be directly applied in unsupervised domain adaptation and any-shot domain adaptation, by further augmenting the available samples in target domain. 
\begin{figure}
    \centering
    \includegraphics[width=0.4\textwidth]{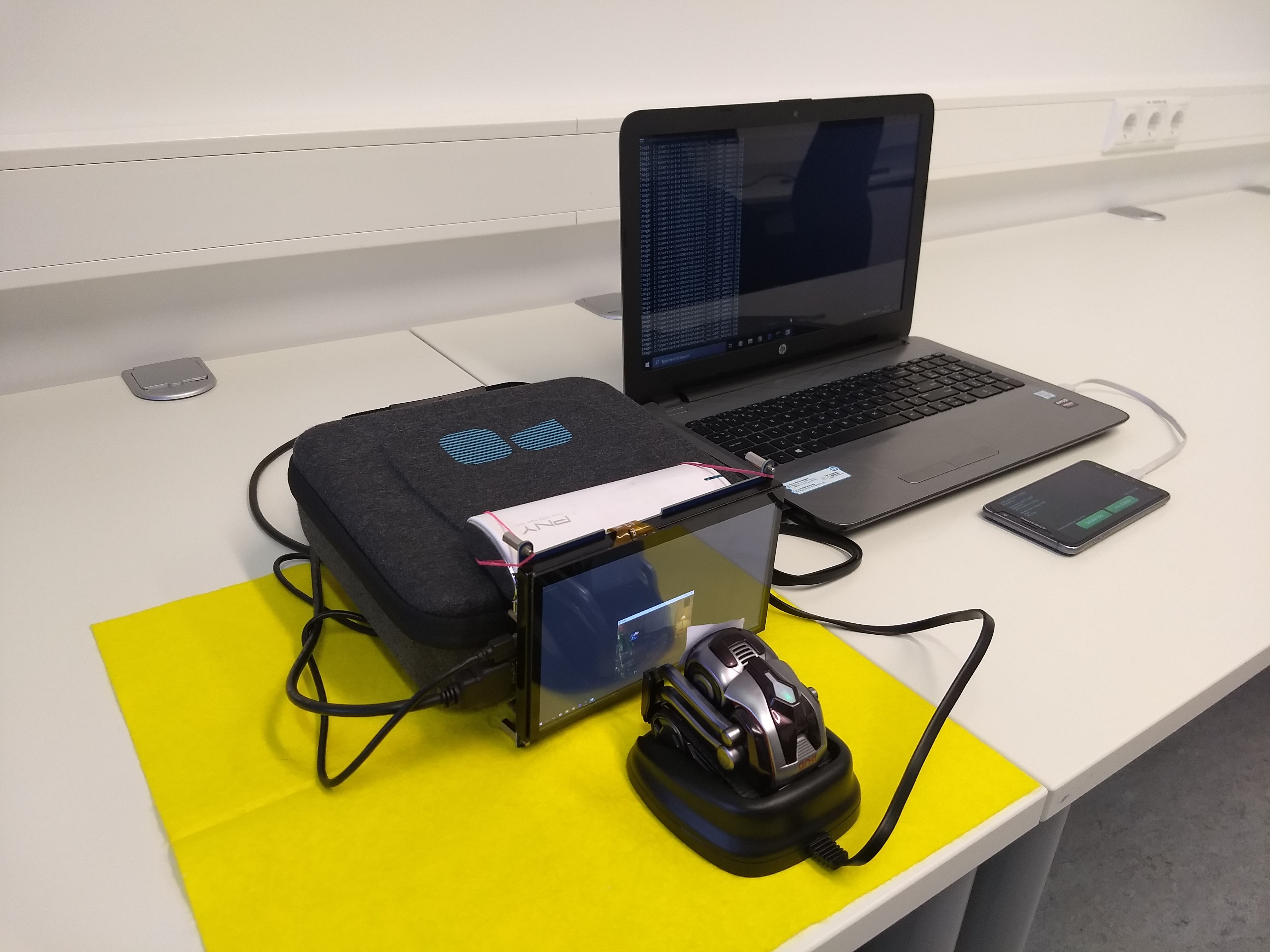}
    \caption{Illustration of our setup for recording pairs of corresponding high and low resolution image: High resolution images are displayed on the small screen,  which is captured by the the built-in camera of the robot.}
    \label{fig:imageRecording}
\end{figure}

The task of mapping an image from one representation to another is generally referred to as an image-to-image translation problem, which can include more complicated mappings such as mapping edges to natural images, colorizing a gray scale image, maps to photos etc. For such tasks, a simple convolutional regression network is not sufficient. Instead, it requires bigger and more powerful architectures employing GANs such as~\cite{Isola16}. Similarly, GAN based approach~\cite{cycleGAN} solves image to image translation when paired data is not available. While \cite{Isola16} or \cite{cycleGAN} have demonstrated impressive results in very complex image-to-image translation tasks, such generative adversarial approach are highly non-trivial to train. Murez~\emph{et al.} ~\cite{murez2018image} employ an image-to-image translation network for domain adaptation with unpaired and unlabeled target domain data using a cycle GAN~\cite{cycleGAN} with additional networks and losses.

As our recording setup does provide us with paired data, we can take the much simpler approach to train a regression network and avoid the (often cumbersome) training of a GAN. 
%However, we do not consider such an approach, as mapping high quality image to lower quality from paired data is a simpler task, and we observed that a simple convolutional regression network is sufficient to do this reasonably well.

\graphicspath{ {./Images/} }
\begin{figure}
\vspace{5pt}
    \centering
    \includegraphics[width=0.5\textwidth]{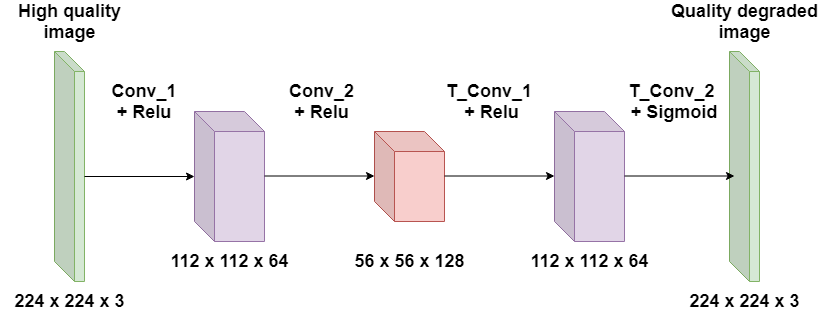}
    \caption{Overview of the proposed domain shifting network}
    \label{fig:BlockDia}
\end{figure}

\section{Proposed Approach}
\subsection{Recording Training Data}
\label{sec:recording_training_data}
Instead of mapping to domain-invariant features as done in many domain adaptation approaches, we create a simple image-to-image network to simulate a low quality image from high quality input image. Our idea is to simplify the domain adaptation problem by creating a dataset of \textit{corresponding} high and low quality images and training a \textit{simple domain shifting network} to map high to low quality images. For this purpose, we exploit the experimental setup shown in Fig.~\ref{fig:imageRecording}: A set of high resolution images is shown on a small screen which are recorded by the robot standing right in front of it. While this of course reduces the set of recorded training images to those in which the low quality image was taken of a screen, i.e., of a luminescent instead of a reflecting object, we will demonstrate that such a setting is sufficient to obtain improved classification accuracy even for real images taken with the low-cost camera. We'd like to point out that the pairs of corresponding images are fed into the domain shifting network (to be described in the next subsection) without any additional registration such that one has to expect small misalignment. This misalignment, however, can be reduced to a minimum by carefully positioning of the robot in front of the screen (as illustrated in Fig.~\ref{fig:teaser}), and the remaining difference does not seem to harm the accuracy of our approach. 

\subsection{A Domain Shifting Neural Network}\label{regress_net}
To map high quality images to low quality images, a domain shifting network is proposed. %Autoencoder in a nutshell is a neural network that is trained for copying input to its output\cite{textbook}. Autoencoders may be viewed as combination of two networks: an encoder part which translates image from image space to latent space and a decoder part which reconstructs the original image from the latent space representation. 
%Autoencoder in past have been used for the purpose of denoising \cite{denoising_AE1}\cite{denoising_AE2}. This variant of autoencoder is named appropriately as denoising autoencoders. For training a denoising autoencoders, a noisy image is feed into the network and the reconstructed image is compared with corresponding clean image.  In this manner autoencoder learns to neglect the noise in a given image while reconstruction. Inspired by this architecture, we propose a network with an intention to add noise to a clean image, to mimic the low-quality camera image of a Cozmo robot. 
This is a convolutional regression network whose input is a high quality image from standard dataset. The network is trained to mimic the corresponding low-quality camera image of a Cozmo robot by minimizing the reconstruction error ($L_2$ loss) between the network output and the corresponding low quality image. Once trained, this network provides a simple way to generate realistic low quality training samples even from previously unseen categories of high quality images.
\begin{figure}
    \centering
    \vspace{.25cm}
    \begin{tabular}{lll}
        \includegraphics[width=2.5cm, height=2.5cm]{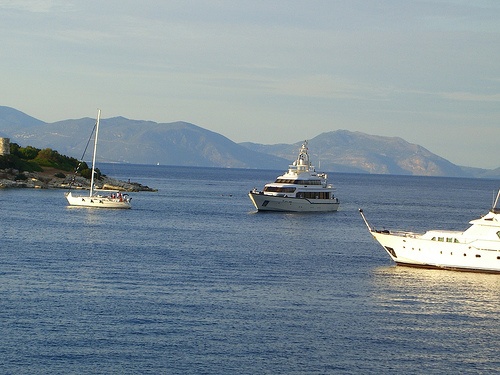}
        & \includegraphics[width=2.5cm, height=2.5cm]{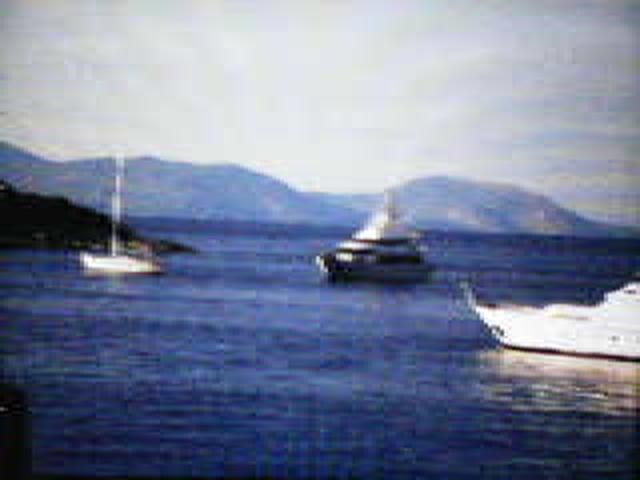} 
        & \includegraphics[width=2.5cm, height=2.5cm]{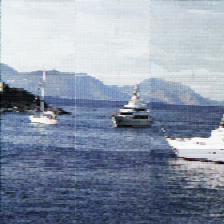} \\
            \includegraphics[width=2.5cm, height=2.5cm]{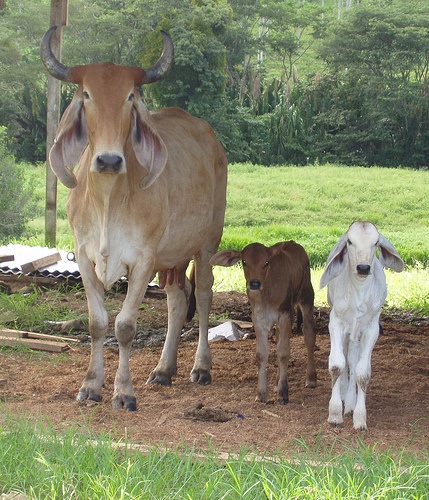}
        & \includegraphics[width=2.5cm, height=2.5cm]{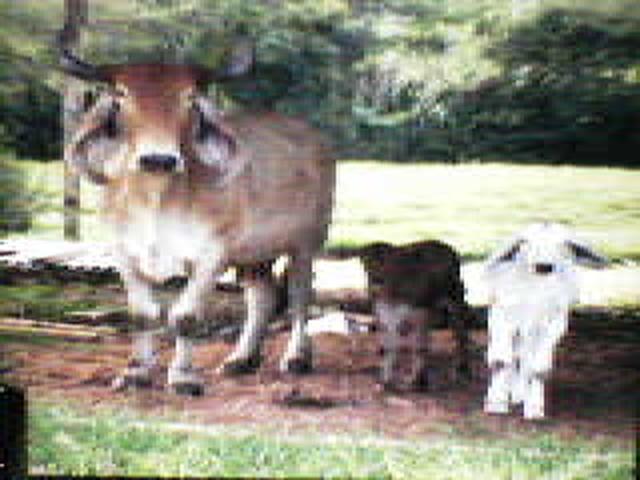}
        & \includegraphics[width=2.5cm, height=2.5cm]{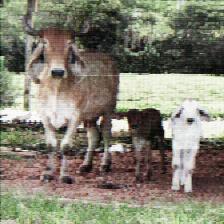} \\
            \includegraphics[width=2.5cm, height=2.5cm]{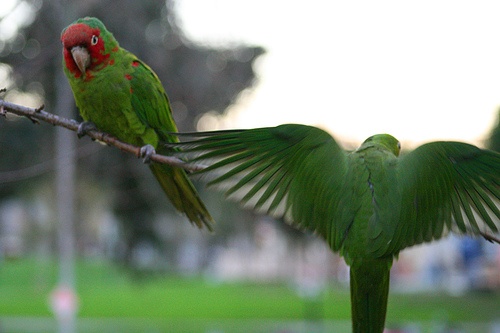}
        & \includegraphics[width=2.5cm, height=2.5cm]{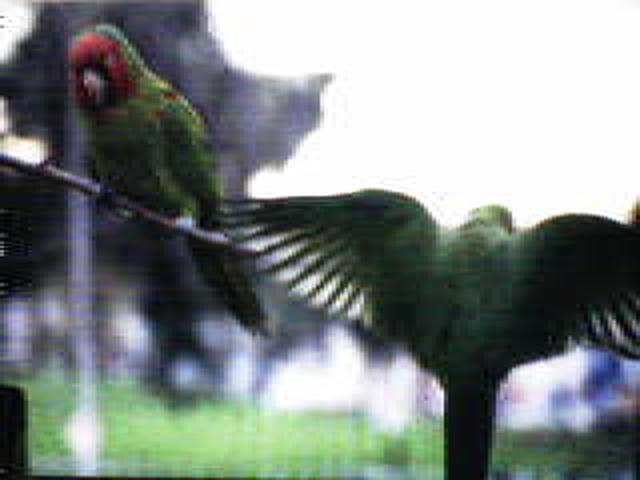}
        & \includegraphics[width=2.5cm, height=2.5cm]{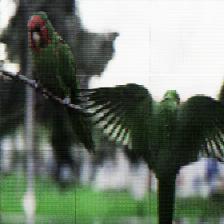} \\            
        \small{a)~Standard}
        & \small{b)~Cozmo recorded}
        & \small{c)~Network output}\\
        %\includegraphics[width=2.5cm,height=2.5cm]{dog.2_Stanford.jpg}
        %\includegraphics[width=2.5cm, height=2.5cm]{dog.2_Cozmo}
        %\includegraphics[width=2.5cm, height=2.5cm]{dog.2_AE}
        %(a) Original  & (b) Degraded & Cozmo 
    \end{tabular}
    \caption{Images from  Pascal VOC \cite{PascalVOC} as illustrated in the left. Corresponding images recorded from Cozmo camera are in the middle, which introduce considerable  distortion in terms of the dynamic range and color reproduction. The images in the right are outputs from the trained domain shifting network, with standard images as input.}
    \label{fig:recording}
\end{figure}

\textit{Architecture}:~The network has a simple 2D convolutional network with 4 convolutional layers, as given below:
\\\\
~$C_{3\to 64}^{3}\downarrow_{2}\to C_{64\to 128}^{3}\downarrow_{2}\to C_{128\to 64}^{2}\uparrow_{2}\to C_{64\to 3}^{2}\uparrow_{2}$
\\\\where $C_{a\to b}^{c}\downarrow_{s}$ represents convolution filter mapping from channel dimension of \textit{a} to \textit{b} and filter size of \text{c} and stride \textit{s}. $C_{a\to b}^{c}\uparrow_{s}$ is a fractional strided convolution~(transpose convolution) filter mapping from a channel dimension of \textit{a} to \textit{b} and using a filter size of \text{c} and stride \textit{s}. This architecture is illustrated in Fig.~\ref{fig:BlockDia}, with the feature map dimension. The first $3$ convolution layers are followed by a rectified linear units (ReLU) as a non-linearity, and we used a sigmoid after the last convolution layer. We do not use any batch normalization.
%encoder section consisted of two 2d-convolutional layer ($3\times 3$ Kernel with padding = 1, stride = 2 and channel = 64(first convolutional),128(second convolutional)), each followed by a relu layer. The decoder section consisted of two 2d-transpose convolutional layer ($2\times 2$ Kernel with padding = 0, stride = 2 and channel = 128(first transpose convolutional),64(second transpose convolutional)). The first transpose convolution layer is followed by a relu layer and the second transpose convolution layer is followed by a sigmoid layer.
\subsection{Simple Domain Adaptation}
In case a labeled target domain dataset for the desired task~(here classification of specific classes) is not available, we use the trained domain-shifting network to map the labeled source domain data to the target domain. Fig.~\ref{fig:recording} shows the mapping from source images to the (lower quality) target domain using our domain shifting network on three examples of our validation data set. We can observe that distortions specific to the target domain in terms of the dynamic range and color reproduction are captured in the network output. This synthetic data along with source domain data is subsequently used to train a network in target domain on the relevant task~(we consider classification).  In this work, we use MobileNet~\cite{MobileNetV2}, a light weight network architecture based on inverted residual blocks containing depthwise separable convolutions, with thin bottleneck layers as inputs and outputs of these blocks. This network architecture is very effective in a variety of computer vision tasks including image classification, object detection and semantic segmentation.

For domain adaptation, we train MobileNet for image classification using clean data together with synthetic data generated by our domain-shifting regression network. This way, we obtain domain invariant features from the network without explicitly trying to minimize the divergence between the features of source and target domain data.  We consider two settings:
\begin{itemize}
    \item The first setting we consider is zero-shot domain adaptation, where our approach does not see any target domain data which is useful in the relevant task. In this case, the unlabeled data used for training the domain shifting network does not contain the categories of images from the relevant task. 
    \item The second setting we consider is unsupervised domain adaptation, in which our regression network is provided with images, which also include a small subset of images containing objects of the final categories of interest.
\end{itemize}
In both cases, the regression network is trained without label information (purely on image-to-image mappings) and subsequently provides synthetic data used to train the classifier network.
%    To also illustrate the effect of having unlabeled target domain data containing the categories of interest, we also train the domain shifting network on the full Pascal VOC dataset, along with corresponding Cozmo captured images.
 
\section{Experiment Setup}
\subsection{Datasets}
For training the domain shifting network, we use images from Pascal VOC dataset~\cite{PascalVOC}. Pascal VOC dataset has a total of $17,125$ images in 20 classes for training and testing. For training the classifier network, and evaluating domain adaptation we use images from the Asirra dataset~\cite{asirra} and a subset of the Imagenet~\cite{ILSVRC15} dataset. The Asirra dataset has $18,697$ images of cats and dogs, with a training-test split of $80:20$. From the Imagenet dataset, we collect images to obtain $5$ classes \{cat, dog, cow, horse and sheep\}, with approximately $2500$ images in each of this classes\footnote{We form $5$ classes by grouping images from Imagenet classes: \{cat, alley cat, Burmese cat, domestic cat\}, \{cow, dairy cattle\}, \{dog, Australian terrier, golden retriever, hunting dog, Labrador retriever\}, \{draft horse, farm horse, horse, male horse, racehorse, wild horse\}, \{black sheep, domestic sheep, sheep, wild sheep\}.} We split this dataset into training, validation and test sets in the ratio $60:20:20$. The images from all these datasets are also recorded by Cozmo robot placed in front of the screen in a dark room as described in section~\ref{sec:recording_training_data}.  Cozmo recording took 1.189 seconds per image. This required around $4$ hours for capturing Pascal VOC dataset, $6$ hours for capturing Asirra dataset and $2.5$ hours for recording the subset of Imagenet. 

\begin{figure}
    \centering
    \begin{tabular}{lll}
        \includegraphics[width=2.5cm, height=2.5cm]{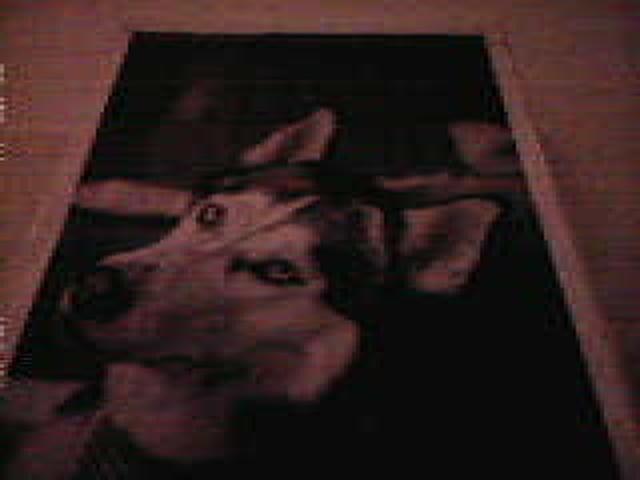}
        & \includegraphics[width=2.5cm, height=2.5cm]{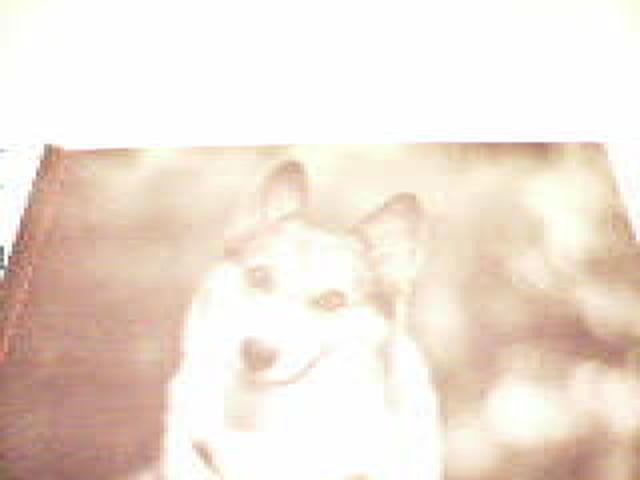} 
        & \includegraphics[width=2.5cm, height=2.5cm]{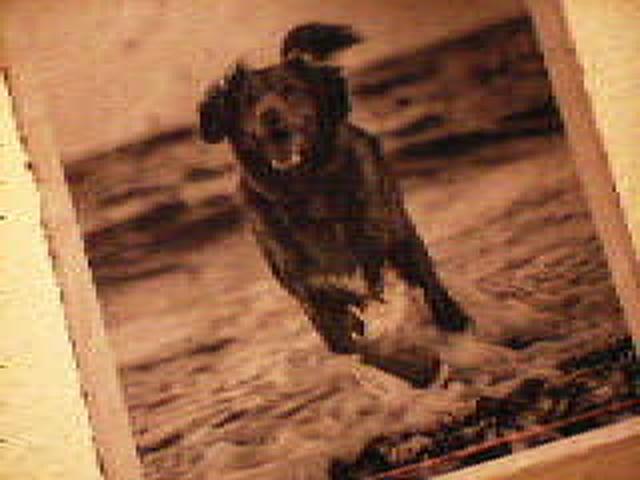} \\
            \includegraphics[width=2.5cm, height=2.5cm]{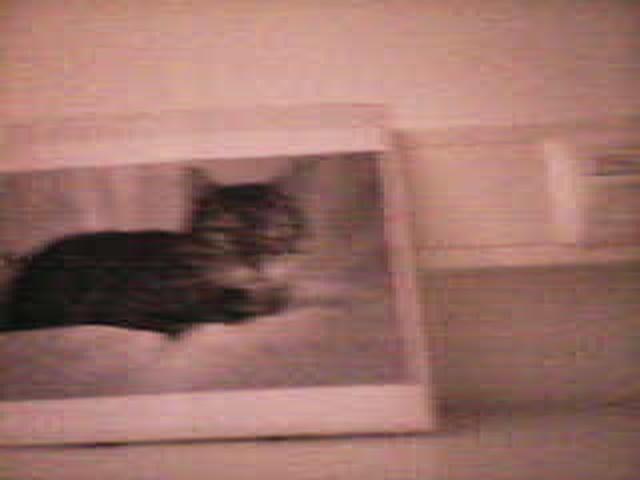}
        & \includegraphics[width=2.5cm, height=2.5cm]{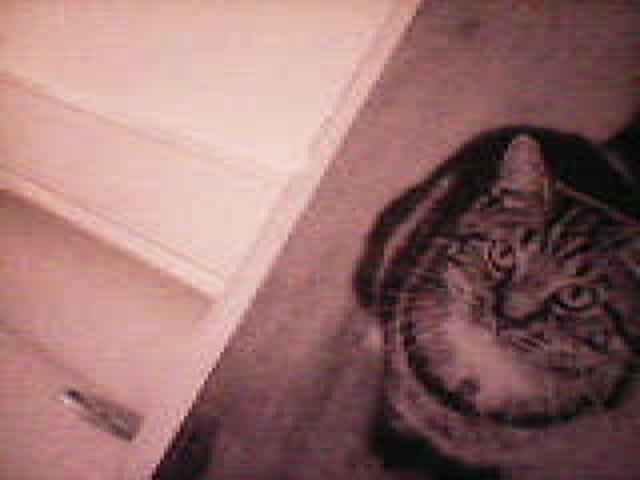}
        & \includegraphics[width=2.5cm, height=2.5cm]{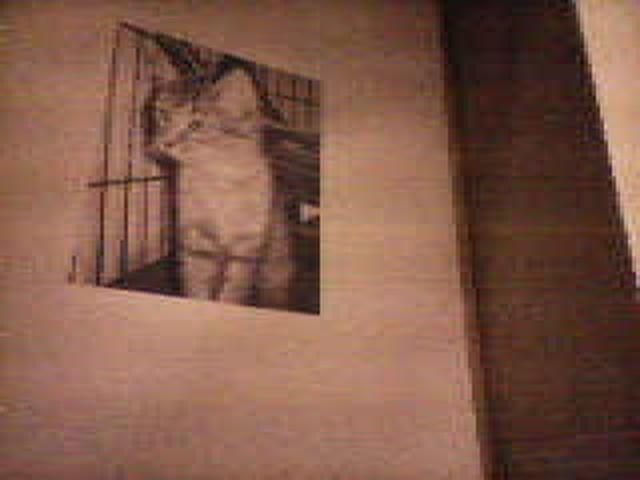} \\

    \end{tabular}
    \caption{Sample images from ``Cozmo in wild''.}
    \vspace*{-5mm}
    \label{fig:cozmowild}
\end{figure}

  We consider two tasks of interest: \textit{i)}~a two-way classification between $2$ classes, cats and dogs, and \textit{ii)}~the classification into our $5$ classes \{cat, dog, cow, horse and sheep\}. For two-way classification~(zero-shot), the domain shifting network is trained on the $18$ remaining classes of the Pascal VOC dataset along with the corresponding low quality images recorded by Cozmo robot. High quality cats and dogs images from the Asirra training set are then mapped to their corresponding low-quality versions by the domain shifting network, which is used to train the classifier network. The evaluation is performed on the low-quality Cozmo captured test images from Asirra dataset. Furthermore, we printed images of $4$ different dogs and $6$ different cats on a paper and captured $233$ images of these cats and dogs under varying illuminations using the Cozmo camera placed at different distances and orientations. We call this setting ``Cozmo in wild'', which is used only in the evaluation. Some samples from this setting are shown in  Fig.~\ref{fig:cozmowild}. 
  
  For the $5-$way classification, we again exclude images from the $5$ classes of interest from the Pascal VOC dataset, and use the remaining ones for training the domain shifting network along with the corresponding low quality Cozmo recorded images. Subsequently, the domain shifting network is applied to the $5$ classes we formed from the Imagenet training set to yield lower quality images, which are subsequently used to train the classification network. 
  
  Finally, we refer to the corresponding \textit{unsupervised domain adaptation} approaches by training the regression network on the full Pascal VOC dataset and not leaving the classes of interest out. The training and evaluation of the classification network, however, remains identical to the zero-shot case.
  
\graphicspath{ {./Images/} }
\vspace{3pt}

\subsection{Baselines}
For comparisons, we obtain the performance references for the fully supervised classification in the source (clean image) domain, and the target (Cozmo captured) domain. We train fully supervised classifiers for both domains by retraining an Imagenet-pretrained MobileNet V2 for the $2$ and $5$ classes respectively. 

Additionally, we compared to the zero-shot domain of ~\cite{peng2018zero}, but unfortunately were not able to even beat the naive supervised training on the source images (even not when pretraining on the latter). Thus, we decided to leave out the specific accuracies of this approach in our numerical results below. Moreover, we tried to compare to the adversarial domain adaption approach in \cite{ganin2014unsupervised}, for which code is provided at \url{https://github.com/jvanvugt/pytorch-domain-adaptation} for an MNIST classification example. However, to show a fair comparison we needed to adapt the classifier to the same network architecture we used, i.e., MobileNet-v2, which subsequently required an adaptation of the discriminator (which otherwise was too weak). Unfortunately, we, again, were unable to find a suitable architecture that improved the results of the naive supervised training on the source domain. While we do believe that adversarial domain adaptation techniques can be very powerful, our experiments demonstrate that balancing the two players in the adversarial training can be a difficult task. More specifically, we did not manage to reach a Nash-equilibrium our adapted discriminator did not manage to decrease the loss. 

%For zero-shot domain adaptation task we compare with the technique in ~\cite{peng2018zero}, since this approach is closest to ours in terms of assumptions and problem addressed.  We use the same architecture for this baseline also, so that the comparison with our approach is fair and not influenced by the choice of architecture. 

%However, to apply ~\cite{peng2018zero}, we need to assume that the source-target domain data pairs for the irrelevant task are labeled, in contrast to unlabeled data used in our approach. Since available data for the irrelevant task is small (about 13,500 images in 15 classes), to obtain the source domain network, we first finetune an Imagenet-pretrained MobileNet V2 on the source data only.  Subsequently for domain adaptation, we follow the two-step procedure of~\cite{peng2018zero}.  We train the target network to minimize the $L_2$ distance between features of target domain images~(Cozmo captured Pascal VOC images from 15 categories) send through the target network and features of the corresponding paired source images send through the source domain CNN. Subsequently, the source domain CNN and classifier are retrained on the relevant source domain images~(5 classes), while simultaneously minimizing the $L_2$ distance between source and target network features on the irrelevant task.

\subsection{Training Details}
We used Pytorch 1.1.0 and Python 3.6.9 for all the experiments. We have made our code  available at \url{https://github.com/Guru-Uni-siegen/Domain-Shifting-Network}. We now describe the details of training for both the domain shifting network and the classification network.
\subsubsection{Domain Shifting Network} %We train the domain shifting network with architecture as described in section~\ref{regress_net} on Pascal VOC dataset and the corresponding Cozmo recorded images, excluding the classes used for classifier. We minimize $L_2$ loss between and clean and Cozmo recorded images for training. 
For training the domain shifting network, we use the Adam optimizer~\cite{kingma2014adam} with $\beta_1= 0.9$ and $\beta_2=0.999$, with an initial learning rate of $0.01$, which is  decreased by a factor of 0.5 every 30 epochs, and train for a total of 100 epochs using a batch size of 32. 

\subsubsection{Classification Networks}
We train all classification networks using stochastic gradient descent with cyclical learning rate scheduling~\cite{smith2017cyclical} with learning rate increasing exponentially from  $1e-5$ to $1e-3$ in 20 steps. We use a batch size of $32$ and train for $100$ epochs and select the model with the best validation error for testing. We start our training with a MobileNet-v2 pretrained on Imagenet and freeze the weights of the first $100$ out of the total number of $157$ layers. %During training we exploit random resizing and cropping, random rotations of up to $10^o$, and random horizontal flips for data augmentation.

% We train on Asirra dataset for two-fold classification and the subset of Imagenet for five fold classification, in fully supervised fashion for the fully supervised baseline on clean images. We also train on the corresponding Cozmo generated images of these datasets for supervised reference baseline for Cozmo images. When we use data augmentation, we use the following transformations on the data: random resized crop, random affine transformation with rotation range of $10^o$, and random horizontal flips.  We employ MobileNet-v2 as our classification network, which has total $157$ layers.  We use an Imagenet-pretrained MobileNet-v2, and freeze the first $13$ features, which correspond to first $100$ layers of the network, and train the remaining layers for both fully supervised baselines and for domain adaptation. For the task of domain adaptation, we use the images generated by domain shifting network for training the classification network.  For all the classification networks, we use stochastic gradient descent for optimization with cyclical learning rate scheduling~\cite{smith2017cyclical} with learning rate increasing exponentially from  $1e-5$ to $1e-3$ in 20 steps. We use a batch size of $32$ and train for $100$ epochs and select the model with the best validation error for testing. 
 
\section{Results}\label{sec:results}

\begin{table}\label{2_way_result}
    \centering
    {\small
    \vspace{5pt}
\begin{tabular}{|l|l|l|l|}
     \hline
     Approach& Standard & Cozmo & Cozmo\\
     &&&in wild\\
    \hline
    Source Supervised& 97.86\% & 86.97\% & 90.13\%\\
    %Source Supervised&\% & \% & \%\\
    %Cozmo Supervised& \% & \% & \%\\
    Ours Unsupervised & 98.76\% & 94.67\% & 91.27\%\\%89.27
   % Ours Unsupervised(Aug) & 98.731\% & 93.717\% & 93.133\%\\
    Ours zero-shot & 98.60\% & 94.24\%  & 95.28 \% \\
   % Ours zero-shot(Aug) & 98.699 \% & 93.860 \% & 90.129 \% \\
    %\hdashline
      Cozmo Supervised& 97.40\% & 95.00\% & 92.27\%\\
    \hline
\end{tabular}
\vspace{1pt}
\caption{Performance comparison for $2-$way classification. The reported numbers are classification accuracy}}
\end{table}
\begin{table}
    \centering 
    {\small
    \vspace{5pt}
\begin{tabular}{|l|l|l|}
    \hline
     Approach& Standard & Cozmo \\
    \hline
    %Standard Augmented & 93.024\% & 75.607\% \\
    %Cozmo Augmented & 84.373\% & 78.935\% \\
    Source Supervised  & 92.87\% & 73.49\% \\
    Ours Unsupervised & 91.66\% & 77.56\% \\
    %Ours Unsupervised(Aug)  & 92.323\% & 80.149\% \\
    Ours zero-shot & 92.09\% & 76.39\% \\
    %Ours zero-shot(Aug) & 91.855 & 79.640\\
    %ZSDDA~\cite{peng2018zero}&&\\
    %\hdashline
    Cozmo Supervised & 84.88\% & 80.15\% \\
    \hline
\end{tabular}
\vspace{1pt}
\caption{Performance comparison for $5-$way classification. The reported numbers are classification accuracy}}
  \label{5_way_result}
\end{table}

A numerical comparison of both the $2$- and $5$-classification problems with the baseline of training the same network on the source data only is given in Tables I and II, respectively. For comparison purposes, we also include the oracle network that is trained on the cozmo-recorded data directly (which we assume to be unavailable). 
%\MM{Mention here which approaches you tried numerically but which failed to give reasonable results.}
%We now investigate the effectiveness of our domain shifting network in domain adaptation, without labeled target domain data. For this we compare our approach with baselines on $2-$way and $5-$way classification. The corresponding results are summarized in Tab.~\ref{tab:2_way} and Tab.~\ref{tab:5_way} respectively, where classification accuracy is compared with the baselines. 

In both, $2-$way and $5-$ way classification, we can observe that networks trained on the source domain do not yield a high classification performance on the target domain. The proposed unsupervised as well as zero-shot domain adaptation techniques improve upon the source supervised network, with our zero-shot approach yielding improvements of $7.27\%$ and $2.9\%$ for the two tasks, respectively. According to the oracle network our performance is nearly optimal for the $2$-class classification and about half-way in between the oracle and naive approach for the $5$-class classification. It is interesting to see that the gain in accuracy on the cozmo images came at no price in accuracy on the higher quality source domain images. 

Comparing the unsupervised and zero-shot approaches, we can see that having paired source and target images of the categories relevant to the classification task does help (showing improvements of $0.43\%$ and $1.17\%$), but does not yield a large margin. 

The general trend of improvement in classification accuracy over standard supervised network baseline is also observed on the `Cozmo in wild' dataset in Table I. However, we also note that this test dataset is small and is not diverse enough to infer the average improvement in performance in this domain.

%The vice-versa also holds true for $5-$ way classification. However, for $2-$ way classification in target domain supervised baseline also performs well on the source images. This could be because of the imagenet pretrained initialization, and subsequent fine-tuning to target domain. Further, $2-$ way classification is an easier task, and we also have more data, about $9000$ images from each class, as opposed to about $2500$ images in each class for $5-$way classification task.

%In both $2-$way and $5-$way classification, we observe that even without real labeled data, we obtain classification performances close to the (oracle) supervised performance without sacrificing accuracy in source domain.  Specifically, our zero-shot approach obtains an improvement of $7.27\%$ and $2.9\%$ in Cozmo domain over the supervised network baseline trained on standard images. Moreover, having unlabeled source-target domain image pairs for the relevant task further improves the classification accuracy in target domain.

%This can be observed in Tab.~\ref{tab:2_way} and Tab.~\ref{tab:5_way} where we find a further improvement of $0.43\%$ and $1.17\%$ in classification accuracy when compared to the zero-shot case. The general trend of improvement in classification accuracy over standard supervised network baseline is also observed on the `Cozmo in wild' dataset in Tab.~\ref{tab:2_way}. However, we also note that this test dataset is small and is not diverse enough to infer the average improvement in performance in this domain.

\section{Discussion}
Domain adaptation addresses the problem of using machine learning algorithms, when there is a shift in the data distribution. The proposed approach presents a practical solution in several real-world applications where the tasks of a robot, or the classes it has to determine, change over time: By recording paired images of high quality and ones recorded by the on-board camera, one can train a domain-shifting network once, and subsequently exploit any online data-base fed through the domain-shifting network to train the robot on new tasks or classes without the need to record new data with the robot. 

Beyond the possibility to avoid the cumbersome labeling of data this way, we'd like to point out that even recording separate images with Cozmo took about 1.19 second per image, rendering the acquisition of gigantic datasets impossible. On the contrary, the forward pass of the domain-shifiting network is in the order of milliseconds.
%Therefore, compared to the original data the system is trained on, there could be unseen classes of images, or there could be changes in the task. It could  be difficult to record new labeled training data, for e.g. when a robot is remotely deployed. In such cases, our introduced domain shifting network may be used. Further, recording and labelling new data is a time consuming process. In our experiments, recording each image with Cozmo took about 1.19 second, where as the synthetic data generation is in the order of milliseconds. Thus the proposed approach is particularly attractive when high amounts of target domain data is needed.

Instead of explicitly minimizing divergence between source and target domain data distributions~\cite{sun2016deep, rozantsev2018beyond, lee2019sliced} or minimizing the distance between features of images from both domains~\cite{peng2018zero}, our approach implicitly learns invariant representation by training on both source and target domain for the classification task. This has the advantage that the performance in the target domain improves significantly, while maintaining good performance in the source domain. A small caveat, however, is the increase in computational load during training due to the increase in training data (considering the union of source and simulated target domain data).

\section{Conclusion}
We proposed a framework to map high quality images to corresponding low quality version using a simple convolutional regression network. This can be used to efficiently generate low quality images of previously unseen categories. We propose a simple domain adaptation approach where we utilize such synthetic data when real labeled data in target domain is not available. Our experiments demonstrate the merit of our simple approach, showing an improved accuracy on the target domain at no sacrifice of source domain accuracy. 
\comment{\begin{figure}
    \centering
    \includegraphics[width=0.5\textwidth]{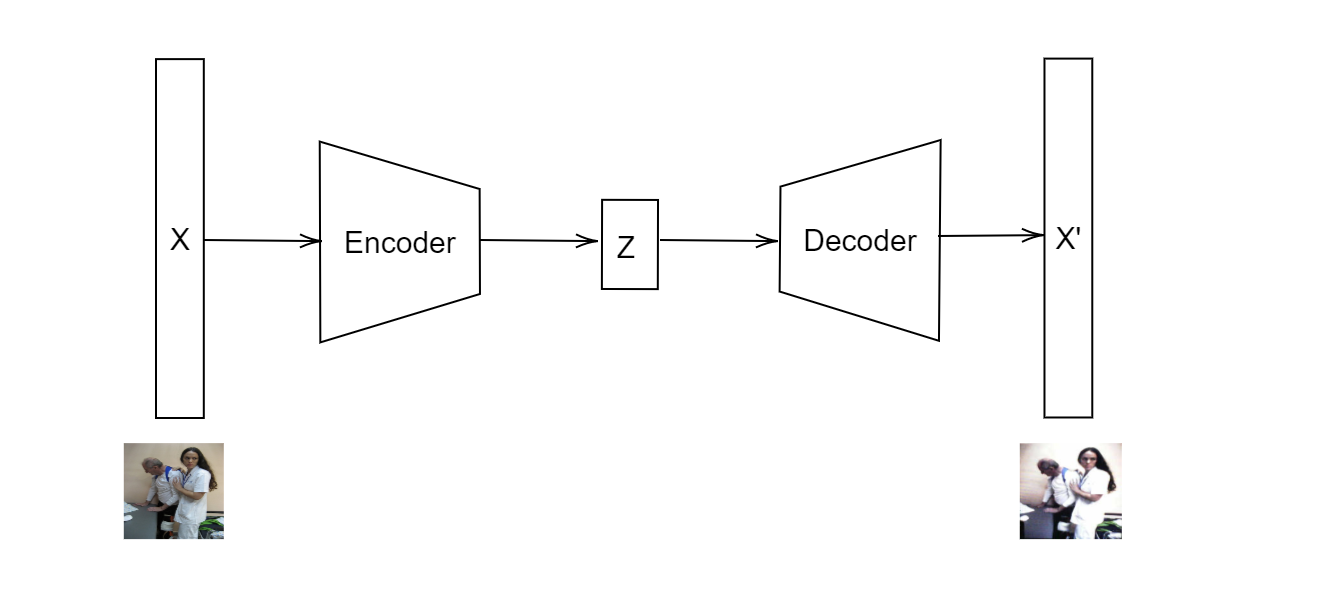}
    \caption{Convolutional autoencoder architecture to train a neural network for degrading high quality image.Input to the encoder is a high quality image and output is the corresponding image taken using low quality camera of the Cozmo robot.}
    \label{fig:AE}
\end{figure}}

%\begin{table}[!t]
%% increase table row spacing, adjust to taste
%\renewcommand{\arraystretch}{1.3}
% if using array.sty, it might be a good idea to tweak the value of
% \extrarowheight as needed to properly center the text within the cells
%\caption{An Example of a Table}
%\label{table_example}
%\centering
%% Some packages, such as MDW tools, offer better commands for making tables
%% than the plain LaTeX2e tabular which is used here.
% Note that the IEEE does not put floats in the very first column
% - or typically anywhere on the first page for that matter. Also,
% in-text middle ("here") positioning is typically not used, but it
% is allowed and encouraged for Computer Society conferences (but
% not Computer Society journals). Most IEEE journals/conferences use
% top floats exclusively.
% Note that, LaTeX2e, unlike IEEE journals/conferences, places
% footnotes above bottom floats. This can be corrected via the
% \fnbelowfloat command of the stfloats package.
%\section*{Acknowledgment}

%The authors would like to thank...

\bibliographystyle{IEEEtran}
\bibliography{refs.bib}
\end{document}